\documentclass{article}

\usepackage[T1]{fontenc}
\usepackage[utf8]{inputenc}
\usepackage{lmodern}
\usepackage{textcomp}

\usepackage{amsmath,amssymb}

\usepackage{graphicx}
\usepackage{svg}
\usepackage{tabularx}
\usepackage{siunitx}
\usepackage{longtable,booktabs,array}

\usepackage{hyperref}
\hypersetup{
  colorlinks=true,
  urlcolor=blue,
  linkcolor=black,
  citecolor=black
}

\usepackage{xurl}
\usepackage{microtype}

\title{A Comparative Study of Open-Source Libraries for Synthetic Tabular Data Generation: \\ SDV vs. SynthCity}
\author{Cristian Del Gobbo}
\date{\today}

\begin{document}

\maketitle

\begin{abstract}
High-quality training data is critical to the performance of machine learning models, particularly Large Language Models (LLMs). However, obtaining real, high-quality data can be challenging, especially for smaller organizations and early-stage startups. Synthetic data generators provide a promising solution by replicating the statistical and structural properties of real data while preserving privacy and scalability. This study evaluates the performance of six tabular synthetic data generators from two widely used open-source libraries: SDV (Gaussian Copula, CTGAN, TVAE) and Synthicity (Bayesian Network, CTGAN, TVAE). Using a real-world dataset from the UCI Machine Learning Repository, comprising energy consumption and environmental variables from Belgium, we simulate a low-data regime by training models on only 1,000 rows. Each generator is then tasked with producing synthetic datasets under two conditions: a 1:1 (1,000 rows) and a 1:10 (10,000 rows) input-output ratio. Evaluation is conducted using two criteria: statistical similarity, measured via classical statistics and distributional metrics; and predictive utility, assessed using a "Train on Synthetic, Test on Real" approach with four regression models. While statistical similarity remained consistent across models in both scenarios, predictive utility declined notably in the 1:10 case. The Bayesian Network from Synthicity achieved the highest fidelity in both scenarios, while TVAE from SDV performed best in predictive tasks under the 1:10 setting. Although no significant performance gap was found between the two libraries, SDV stands out for its superior documentation and ease of use, making it more accessible for practitioners.
\end{abstract}

\vspace{1em}

\section{Introduction}
The performance of machine learning models, particularly in supervised
learning tasks, is heavily dependent on the availability of large
volumes of high-quality data\cite{kariluoto2021quality}. However, real-world datasets are
often limited due to privacy concerns, regulatory constraints, high
labeling costs, or data sparsity, especially in domains such as
healthcare, energy systems, and finance\cite{ghassemi2018opportunities}. As a result, synthetic
data generation has emerged as a promising alternative that can
supplement or even replace real data in certain use cases\cite{stoian2025survey}. While
synthetic data has been widely explored in computer vision and natural
language processing\cite{nikolenko2019synthetic, li2021data}, its application to structured, tabular
data, common in many real-world business and scientific settings, has
gained traction more recently. To support this growing need, several
open-source tools and libraries have been developed. Among the most
prominent are the Synthetic Data Vault (SDV) and Synthicity, both of
which provide user-friendly APIs for generating synthetic tabular data
using a variety of models, ranging from classical probabilistic
approaches to modern deep generative networks.

Despite the growing popularity of these tools, few studies have
systematically compared their performance under controlled, real-world
conditions. This study aims to fill that gap by evaluating and comparing
SDV and Synthicity in terms of both statistical fidelity and downstream
predictive utility. Using a real-world energy consumption dataset, we
generate synthetic data under two experimental conditions: one where the
output data matches the input size (1:1), and another where it scales
tenfold (1:10). We assess performance using both statistical similarity
metrics and a Train-on-Synthetic, Test-on-Real (TSTR) predictive
evaluation pipeline.

Our findings offer practical insights into the strengths and limitations
of each library and provide guidance for researchers and practitioners
selecting synthetic data generation tools for tabular data tasks.

\section{Literature Review}

In the past few years, Large Language Models (LLMs) have become
prominent, and with the release of commercial models like ChatGPT by
OpenAI in November 2022\cite{openai2022chatgpt}, their power became available to anyone
with internet access, greatly impacting many aspects of daily life\cite{george2023chatgpt}.

A common belief behind the success of LLMs is the scaling law of
computing, model size, and, perhaps most importantly, the high quality
of pre-training data\cite{chen2024diversity}. The biggest LLMs today are often
pre-trained on trillions of tokens. For example, GPT‑3 was famously
trained on nearly 500 billion tokens\cite{brown2020language}, from a mixture of web
text, books, and other sources. GPT‑4 is rumored to have used well over
a trillion tokens\cite{openai2023gpt4}, and Anthropic's Claude reportedly relies on
a similarly large-scale corpus, likely in the hundreds of billions to
trillions of tokens\cite{bai2022constitutional}. The exact numbers for GPT‑4 and Claude
have not been officially disclosed by OpenAI and Anthropic, but external
analyses report similar figures\cite{bubeck2023sparks}.

However, acquiring such a massive quantity of high-quality data has
become more challenging\cite{villalobos2022will}. Many sources are now gated behind
paywalls, restricted by copyright, or filtered due to data quality
concerns\cite{perelkiewicz2024review}. As the demand for high-quality training data grows,
finding scalable solutions for future LLM development remains an open
question.

As a remedy, synthetic data has been widely adopted in training LLMs,
offering a more accessible and controllable alternative to real-world
data\cite{bauer2021identify, liu2024best}. Chen et al.\cite{chen2024diversity} conducted a study on the
measurement of diversity in synthetic data and its impact on LLM
performance. They examined how synthetic data diversity influences both
pre-training and fine-tuning stages, introducing a new diversity metric
called LLM Cluster-Agent, designed specifically to evaluate the
diversity of synthetic datasets. They define LLM Cluster-Agent as ``a
diversity measure pipeline that leverages LLM's ability to interpret
semantic meanings and understand rich contexts of text samples for
clustering''. This metric is particularly suited for text-based
synthetic data, which is commonly used in the pre-training process of
large LLMs, rather than for tabular data. Through a series of controlled
experiments with 350M and 1.4B parameter models, Chen et al.
demonstrated that higher diversity in synthetic data correlates
positively with both pre-training and fine-tuning performance.
Interestingly, their findings suggest that synthetic data diversity in
pre-training has an even stronger effect on fine-tuning than on
pre-training itself. Although this study differs from our goal of
comparing tabular data generators rather than synthetic text data, it is
still relevant because it highlights how synthetic data can be
effectively leveraged in the pre-training of LLMs.

The use of synthetic data generators for training LLMs, however, is not
their only application. In fact, synthetic data generation is now widely
used across multiple domains. Lu et al.\cite{lu2024machine} presented a
comprehensive review of existing studies on employing machine learning
for synthetic data generation, highlighting applications spanning
computer vision, speech, natural language processing, healthcare, and
business domains. Their review categorizes existing approaches based on
machine learning techniques, with a particular emphasis on deep
generative models, including GANs, VAEs, and reinforcement
learning-based methods. One of the key findings of their study is that
the effectiveness of synthetic data depends on the application domain.
In computer vision, synthetic datasets are frequently used to train
models for object detection, facial recognition, and domain adaptation
when real-world labeled data is scarce. In speech processing, synthetic
data has proven valuable in speech synthesis and voice cloning, reducing
the need for extensive manually labeled datasets. In natural language
processing, it is used to augment training datasets for tasks such as
language modeling and machine translation. In healthcare, synthetic data
generation enables the use of privacy-preserving patient data,
facilitating medical research and predictive modeling without
compromising sensitive information. In business and finance, synthetic
data is used to simulate market behaviors, detect fraudulent
transactions, and improve risk assessment models. Beyond applications,
Lu et al. also discuss key challenges in synthetic data generation,
particularly issues of data fidelity and bias. They emphasize that while
synthetic data can approximate real-world distributions, its utility
depends on the balance between realism and generalization. Their study
provides an important foundation for understanding the broad
applicability of synthetic data generation, reinforcing its relevance
across various fields where data limitations exist.

Since this study compares existing open-source Python packages for
synthetic data generation, it is essential to review the technical
aspects and identify the most suitable and widely used models. Various
approaches exist for generating synthetic data, ranging from graph-based
models and probabilistic methods to deep neural networks. To highlight
some of the most well-known models, I refer to the work of Bauer et al.\cite{bauer2024comprehensive}, which provides a comprehensive overview of synthetic data
generation techniques. This section will not delve into the technical
details of each model, as a more precise definition of the models used
in the comparison will be presented in the Method section.

Starting with probabilistic and statistical models, one of the most
widely implemented is the Gaussian Mixture Model (GMM). GMMs are density
estimation algorithms primarily used for clustering, but they can also
serve as generative probabilistic models. They are commonly applied to
tabular data and time-series generation. A GMM consists of N Gaussian
distributions, each representing a continuous, symmetric probability
distribution. Another important probabilistic model is the Markov Chain,
which is used for generating sequential data by modeling infinite
sequences of symbols where the probability of each symbol depends only
on the previous n symbols. These models are widely applied in text
generation and time-series synthesis.

Bayesian Networks (BNs) offer a graphical approach to modeling
dependencies between variables. They are structured as Directed Acyclic
Graphs (DAGs), where nodes represent random variables, and edges define
their conditional dependencies. Each variable follows either a
continuous or a discrete probability distribution. Synthcity, one of the
Python packages we will analyze, implements Bayesian Networks as they
are particularly effective for structured synthetic data generation,
including privacy-preserving applications. The second Python package in
our study, SDV, utilizes another probabilistic model, the Gaussian
Copula. A copula function represents the joint probability distribution
of a continuous random vector by separating the individual marginal
distributions from the dependency structure between variables. More
details on Bayesian Networks and Gaussian Copulas will be provided in
the Method section.

Even though probabilistic and statistical models are still widely used,
deep learning methods have become the dominant approach for
state-of-the-art synthetic data generation. Among them, one of the most
well-known frameworks is Generative Adversarial Networks (GANs). GANs
consist of two neural networks, a generator (G) that creates synthetic
data from random noise and a discriminator (D) that determines whether a
given sample comes from the generator or the real training data. The
authors of the original GAN paper describe this system as a ``minimax
two-player game'', where the generator continuously improves its ability
to fool the discriminator, while the discriminator becomes better at
distinguishing real from fake data\cite{goodfellow2014generative}. Over time, numerous
variations of the classic GAN architecture, originally implemented with
Multi-Layer Perceptrons (MLPs), have emerged to improve stability,
control, and performance. Deep Convolutional GANs (DCGANs)\cite{radford2015unsupervised}
introduced the use of convolutional layers instead of fully connected
layers, allowing the generator to better capture spatial hierarchies in
data, significantly enhancing the quality of image generation.
Conditional GANs (cGANs)\cite{mirza2014conditional} addressed the uncontrolled nature of
GAN outputs by introducing conditioning variables, such as class labels
or additional attributes, enabling the generator to produce targeted
synthetic samples. Another major advancement came with Wasserstein GANs
(WGANs)\cite{arjovsky2017wasserstein}, which improved training stability by replacing the
traditional Jensen-Shannon divergence with the Wasserstein distance,
mitigating common issues such as mode collapse and leading to more
reliable convergence.

Another widely used deep learning-based approach for synthetic data
generation is Variational Autoencoders (VAEs). VAEs are probabilistic
generative models designed for latent space learning, enabling the
generation of high-dimensional synthetic data such as images and text.
Unlike GANs, which learn to generate data through adversarial training,
VAEs model the data distribution explicitly by encoding inputs into a
latent space and then reconstructing them via a decoder\cite{kingma2013auto}. While
VAEs do not always produce sharper images compared to GANs, they offer
greater control over latent variables, making them useful for tasks
requiring structured and interpretable representations.

For image and text synthesis, powerful generative models are Diffusion
Models\cite{ho2020denoising}. These models operate as Markov chains, where data is
incrementally noised in a forward process over T steps, and the model
learns to reverse this process, gradually denoising the input back to
the original data distribution. Diffusion models have gained attention
for their ability to generate highly detailed images, surpassing GANs in
certain text-to-image tasks.

Perhaps the most influential deep learning model in text synthesis, and
beyond, is the Transformer architecture. First introduced in 2017 in the
seminal paper ``Attention Is All You Need''\cite{vaswani2017attention}, Transformers
gained widespread recognition following the release of LLMs such as
ChatGPT, which are built upon Transformer-based architectures. At their
core, Transformers are sequence-to-sequence transduction models
structured with an encoder-decoder mechanism. Unlike previous recurrent
architectures (RNNs and LSTMs), Transformers allow for full
parallelization, drastically improving efficiency and scalability. The
key innovation behind Transformers is the multi-head self-attention
mechanism, which enables models to capture long-range dependencies in
data with a constant number of sequential operations, rather than the
sequential processing bottleneck of RNNs. This shift allowed
Transformers to excel in language modeling, translation, and generative
tasks, building the foundation for modern LLMs.

To narrow the scope and examine studies similar to this one, as the
conclusion of this literature review, we will analyze research comparing
synthetic data generation techniques in real-world applications. One
such study was conducted by Akiya et al.\cite{akiya2024comparison}, which evaluates
various synthetic data generation methods for control group survival
data in oncology clinical trials. The primary objective of their
research was to determine the most suitable synthetic patient data (SPD)
generation method for oncology trials, focusing specifically on
progression-free survival (PFS) and overall survival (OS), key
evaluation endpoints in clinical oncology. In their study, Akiya et al.
compared four distinct synthetic data generation techniques,
incorporating both probabilistic/statistical methods and deep
learning-based approaches. The traditional methods included
Classification and Regression Trees (CARTs) and Random Forest (RF),
while more complex models consisted of Bayesian Networks (BNs) and
Conditional Tabular Generative Adversarial Networks (CTGANs). To
evaluate performance, the researchers generated 1,000 synthetic datasets
per method and assessed their effectiveness based on both statistical
similarity and visual analysis. The results indicated that traditional
tree-based methods outperformed deep learning-based techniques,
particularly when trained on relatively small datasets, which is common
in clinical trials. CART and RF demonstrated superior performance, with
CART emerging as the most effective method, as its synthetic data
closely matched the statistical properties of real patient survival
data. On the other hand, Bayesian Networks (BNs) and CTGANs did not
perform well, mainly due to their higher data requirements. These models
typically require larger training datasets to learn meaningful patterns
and generate synthetic data that aligns well with real-world statistical
distributions.

While the previous study provides insights into synthetic data
generation techniques, comparisons between specific open-source Python
packages remain scarce. This study aims to fill that gap in the
literature by providing a comparative analysis of two of the most
popular open-source Python libraries for synthetic data generation: SDV
and Synthcity. To evaluate these two packages, we will compare different
models available in each framework. For SDV, we will analyze the
Gaussian Copula synthesizer, Conditional Tabular GAN (CTGAN), and
Tabular Variational Autoencoder (TVAE). For Synthcity, we will evaluate
the Bayesian Network synthesizer, as well as CTGAN and TVAE, to ensure a
direct comparison between the shared models across both packages. The
synthetic data will be generated using a dataset from a publicly
available repository in the UCI Machine Learning repository. It was
collected by continuously monitoring a low-energy house in Belgium for
137 days, capturing both electrical energy consumption and environmental
data. To assess the quality and effectiveness of the generated data, we
will employ two key evaluation metrics:

\begin{enumerate}
\def\labelenumi{\arabic{enumi}.}
\item
  Statistical Difference: Measured by comparing synthetic and real data
  distributions using reliable statistical functions provided by SDV,
  along with custom statistical comparison methods.
\item
  Predictive Utility: Evaluated by training Machine Learning models on
  both real and synthetic data and comparing the performance metrics
  (e.g., accuracy, precision, recall) of the models trained on each
  dataset.
\end{enumerate}

This study aims to identify the best-performing open-source package for
synthetic data generation and provide a comprehensive comparison of the
quality and utility of the data generated by SDV and Synthcity. By doing
so, we hope to contribute valuable insights into the strengths and
limitations of these tools, guiding researchers and practitioners in
selecting the most suitable synthetic data generation framework for
their needs.

\section{Methods}
\subsection{Data Generators Description}
In this research, we\textquotesingle ll use four different Data
Generators with well-known mathematical properties: Gaussian Copula,
Bayesian Networks, Conditional Tabular GAN (CTGAN), and Tabular
Variational Autoencoder (TVAE). The first two are statistical methods,
whereas the latter two are deep neural network techniques.
Let\textquotesingle s first examine the mathematical properties of the
statistical methods.

Gaussian copula methods model the joint distribution of a table by
combining each column's marginal distribution with a copula function
capturing inter-column dependencies. The foundation is Sklar's theorem\cite{jaworski2010copula}, which states that any multivariate distribution
\(F\left( z_{1},\ldots,\ z_{d} \right)\) can be decomposed as
\begin{equation}
F\left( z_{1},\ \ldots,\ z_{d} \right) = C\left( F_{1}\left( z_{1} \right),\ldots,\ F_{d}\left( z_{d} \right) \right),
\end{equation}

where \(F_{i}\left( z_{i} \right)\) are the marginals and \(C\ \)is the
copula describing their dependency structure. A Gaussian copula assumes
this dependence is encoded by a multivariate normal distribution with a
correlation matrix, while allowing arbitrary marginals. In practice, one
first estimates each column's CDF, then transforms the data into a
latent space with uniform or Gaussian marginals. First, you estimate a
CDF \({\widehat{F}}_{i}\) for each column \(i\ \)of the real data. Then,
transform the real data into uniform variables
\(U_{i}\  = \ {\widehat{F}}_{i}(Z_{i})\) and fit a copula
\(\widehat{C}\) to capture their joint dependence, which for a Gaussian
copula means computing the correlation matrix. Lastly, draw a synthetic
sample \(\left( U_{1},\ldots,\ U_{d} \right)\ \sim\ \widehat{C}\) from
the copula model, and invert the transform by applying the inverse CDF
for each column,
\(Z_{i}\  = \ {\widehat{F}}_{i}^{- 1}\left( U_{i} \right).\) This yields
a synthetic data record in the original space.

Patki et al.\cite{patki2016synthetic} introduced the open-source package Synthetic Data
Vault (SDV) by using Gaussian copulas to model tabular data. In their
approach, all columns are converted to a standard normal scale to remove
the effect of each column distribution shape, before estimating the
covariance matrix of the joint Gaussian copula. After modeling the
correlations, synthetic rows are sampled by drawing from the
multivariate normal and then transforming back to each column's domain.
One limitation, however, is that purely categorical fields cannot be
directly handled by the Gaussian copula, since the copula operates in a
continuous space. The SDV work addressed this by encoding categories as
ordinal values in the range [0,1] so that they could be treated like
continuous variables in the copula model. In general, the key advantage
of Gaussian copula models is that they are statistical methods, as
opposed to neural networks, offering a relatively simple mathematical
formulation and a stable fitting procedure, while still capturing
complex dependencies through the copula. However, if the data contains
many discrete variables or highly non-linear dependencies, a copula
model, like the Gaussian copula, might struggle.

Another class of tabular synthesizers uses Bayesian networks (BN) to
learn the joint distribution from the data. A Bayesian network consists
of a directed acyclic graph, in which the nodes represent variables of
the data (columns of a data frame), and a set of conditional probability
distributions for each node given its parent nodes\cite{young2009using}. In essence,
the BN factorizes the joint probability

\begin{equation}
P\left(X_{1}, \ldots, X_{d}\right) = \prod_{i} P\left(X_{i} \mid \text{Parents}(X_{i})\right),
\end{equation}

representing inter-column relationships as conditional dependencies.
Once a BN is learned from the real dataset, generating synthetic data is
simple: one can sample from the network by first sampling the root
nodes, which are variables with no parents according to the learned
graph, and then sampling descendant nodes conditional on their parents'
sampled values, propagating through the network until all variables have
values. This produces a synthetic record that statistically mirrors the
correlation captured in the BN.

An example of Bayesian Network usage can be found in the research by
Zhang et al.\cite{zhang2017privbayes}, which introduces PrivBayes, a method utilizing
Bayesian networks to generate synthetic data under differential privacy
constraints. PrivBayes learns a dependency graph over the attributes
(columns) and then draws synthetic tuples (rows) by sampling that
Bayesian Network. In general, a learned BN can accurately reproduce
multi-variable interactions present in the original data, especially for
mixed categorical data. Therefore, BN have the advantage of being
well-understood models in probability theory, and they inherently ensure
that generated samples are consistent with the conditional distributions
observed in the real data. On the other hand, Bayesian network
generators face some practical challenges. In fact, they often require
discretizing continuous variables or assuming parametric forms for
continuous conditional distributions, which can introduce errors. Moreover, learning the optimal network structure for
high-dimensional data can be computationally expansive and may require
prior knowledge.

On the other hand, compared to statistical models such as the Gaussian
Copula and Bayesian networks, deep learning methods exist for generating
tabular synthetic data. As previously stated, this study will utilize
the Conditional Tabular GAN (CTGAN) and Tabular Variational Autoencoder
(TVAE). CTGAN is a deep generative model specifically designed for
tabular data, introduced by Lei Xu and colleagues in 2019\cite{xu2019modeling}. It
extends the standard Generative Adversarial Network architecture,
consisting of a generator that synthesizes data and a discriminator that
attempts to distinguish between real and synthetic data. CTGAN
introduces novel components to tackle particular challenges of tabular
data, such as mixed data types, imbalanced categories, and complex
distributions. Unlike GAN architectures commonly used in image
generation, which typically rely on convolutional networks\cite{radford2015unsupervised},
CTGAN employs fully connected multilayer perceptrons for both its
generator and discriminator.

Lei Xu et al. introduce three key technical innovations in the CTGAN
framework: Mode-Specific Normalization, a Conditional Generator, and a
Training-by-Sampling strategy. Mode-Specific Normalization refers to
CTGAN's use of variational Gaussian Mixture models to preprocess
continuous columns. Rather than employing simple min-max scaling, as
often seen in classic GAN implementations\cite{zhao2024ctab}, each continuous
column is modeled as a mixture of Gaussians. Each value is normalized
according to the ``mode'' (Gaussian component) it most likely belongs
to. This normalization strategy helps the generator learn multi-modal
and non-Gaussian distributions by providing a richer and more expressive
representation of the continuous data. Essentially, it maps a continuous
variable into a higher-dimensional space (with dimensions corresponding
to mixture components), effectively addressing issues related to
non-Gaussian distributions.

Another significant innovation of CTGAN is the Conditional Generator.
Unlike traditional methods that generate entire rows unconditionally,
CTGAN conditions the generation process on specific discrete column
values during training. For example, if a table has a categorical column
labeled "Education" with categories such as "High School," "Bachelor,"
and "Master," the model may fix a specific category (e.g., "Master")
during certain training iterations and train the generator to produce
rows conditioned on this category. This conditioning is implemented by
appending a one-hot vector to the generator\textquotesingle s input to
indicate the selected category and by filtering real data accordingly
when updating the discriminator. Consequently, the GAN explicitly learns
the conditional distributions \(P(other\ columns\ |\ Category = k)\) for
each category \(k\) within a categorical column, significantly improving
the fidelity of synthetic data for imbalanced categories.

Accompanying the conditional generator, CTGAN incorporates a
Training-by-Sampling strategy to select training minibatches in a
balanced manner. When conditioning on a particular discrete category,
the method selects real data batches exclusively from rows containing
that category to update the discriminator, rather than randomly sampling
from the entire dataset. This targeted sampling approach aligns the
generator's conditioning with the discriminator's data distribution,
preventing the generator from being biased toward majority classes.
Thus, the GAN effectively trains on one sub-population at a time,
implementing a form of oversampling for rare categories.

Regarding its architecture, CTGAN employs the Wasserstein GAN objective
with gradient penalty (WGAN-GP) to ensure stable training\cite{gulrajani2017improved} and
incorporates PacGAN\cite{lin2018pacgan}, which modifies the discriminator to
jointly evaluate multiple samples, thus further mitigating mode
collapse. The architecture combines a noise input z with the
aforementioned condition vector, feeding the generator to produce
synthetic rows. The discriminator then attempts to distinguish between
real and synthetic data. Thanks to Mode-Specific Normalization, the
generator's output for each continuous feature can be accurately mapped
back to the real data space, preserving the original distributions.

In the same paper in which Xu et al.\cite{xu2019modeling} introduced the CTGAN for
the first time, they also presented a the Tabular Variational
Autoencoder (TVAE), a deep generative model that applies Variational
Autoencoder (VAE) framework to tabular data. TVAE adapts a standard
variational autoencoder to handle mixed data types. In a VAE, there are
two networks: an encoder \(q_{\phi}(z\ |\ x)\) that maps an input data
sample \(x\) (a row in the table) to a latent representation \(z\), and
a decoder \(p_{\theta}(x\ |\ z)\) that tries to reconstruct the original
data from the latent code. The model is trained by maximizing the
evidence lower bound (ELBO), which consists of a reconstruction term,
ensuring that the decoded output matches the input data, and a
regularization term pushing the latent \(z\) to follow a specific
standard distribution, usually standard normal\cite{yadav2024rigorous}.

To make VAEs work for tabular data (TVAE) Xu et al. use similar data
pre-processing as for CTGAN, for instance by converting categorical
columns into one-hot vectors and using Mode-Specific Normalization for
continuous columns, so that the encoder and decoder can effectively
model them. The decoder network \(\ p_{\theta}(x\ |\ z)\ \)in TVAE
outputs parameters sufficient to reconstruct each column: for continuous
columns, it might output means/variances (if assuming a Gaussian output
distribution), and for categorical columns, it can output logits for
each category, i.e. softmax probabilities for one-hot output. On the
other hand, the encoder network \(q_{\phi}(z\ |\ x)\) likewise must
handle one-hot inputs for categorical columns and numeric inputs for
continuous columns. With this implementation, TVAE is able to capture
the joint distribution of heterogeneous tabular columns in its latent
space. Once trained, synthetic data generation is done by sampling a
latent vector \(z\) from the prior (e.g. \(\mathcal{N}(0,\ I)\)) and
feeding it into the decoder to produce a new synthetic row. Because the
decoder was trained to produce realistic combinations of values, in
order to match the real data distribution, the sample outputs resemble
real records. Xu et al. also demonstrated that a ``vanilla'' VAE, when
trained with the right loss functions and data encoding, can be
competitive with GAN-based models for tabular data. TVAE's mathematical
formulation is essentially the VAE probabilistic model: it seeks to
maximize

\begin{equation}
\mathbb{E}_{q_{\phi}\left( z\mid x \right)}\lbrack logp_{\theta}(x \mid z)\rbrack - D_{KL}(q_{\phi}(z \mid x) \parallel p(z))
\end{equation}

over the real data, thereby learning \(p_{\theta}(x\ |\ z)\) which can
be sampled. Yadav, Parul, et al.\cite{yadav2024rigorous} concisely summarize TVAE as
``a novel VAE for tabular data using two neural networks (encoder and
decoder) trained with ELBO loss''. In practice, TVAE offers a more
straightforward training process than GANs, i.e. no adversarial game to
balance, and can be easier to converge, though it may require careful
tuning to get the decoder to model categorical distributions accurately.

\subsection{Real Data Brief Description}
The dataset used in this experiment is publicly available from the UCI
Machine Learning Repository. It was collected by continuously monitoring
a low-energy house in Belgium for 137 days, capturing both electrical
energy consumption and environmental conditions. Energy usage data for
various household appliances were recorded every 10 minutes using m-bus
energy meters, while environmental data, including temperature and
humidity, were collected from different rooms via a wireless ZigBee
sensor network. Additionally, meteorological data from a nearby airport
(e.g., temperature, humidity, wind speed, visibility) were integrated
into the dataset based on matching timestamps\cite{candanedo2017data}.

The complete dataset contains 19,735 records and 29 features. However,
for this study, only a subset of these data was used. Specifically, this
subset of real data serves as the initial input to synthetic data
generators. The aim is to evaluate how effectively these generators can
expand and generalize from a limited amount of real data, testing their
ability to create larger synthetic datasets with similar statistical and
predictive characteristics.

\subsection{Comparison Process and Methods}

The goal of this study is to compare the models and tools offered by two
of the most widely used Python packages for synthetic data generation:
SDV and Synthicity. A total of six models were selected, three from each
package, to evaluate and compare their performance. From SDV, the models
used were Gaussian Copula, CTGAN, and TVAE; while for Synthicity, the
models included Bayesian Network, CTGAN, and TVAE. A detailed
description of each model is provided in the preceding section.

To evaluate how well the synthetic data replicates the original data,
two key metrics were used: statistical similarity and predictive
utility. The statistical similarity was assessed using a custom scoring
function that compares the real and synthetic datasets on a
column-by-column basis. The function computes normalized differences in
key statistical properties for numerical columns, namely mean, median,
and standard deviation, as well as two distributional distance measures:
the Kolmogorov--Smirnov (KS) statistic and the Wasserstein distance. For
categorical features, similarity was assessed by comparing the mode of
each column between the real and synthetic data. Each column score was
computed as a weighted combination of these metrics, normalized between
0 and 1, where lower statistical divergence yields higher scores. The
final dataset-level score is the average of all column scores, scaled to
a 0--100 range, with 100 indicating a perfect match between the
synthetic and real datasets.

\begin{figure}[htbp]
  \centering
  \includegraphics[width=1\linewidth]{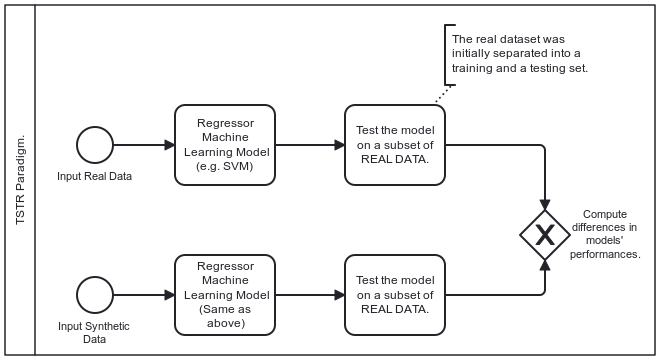}
  \caption{TSTR paradigm explained visually.}
  \label{fig:tstr}
\end{figure}

The predictive utility was assessed using the Train on Synthetic, Test
on Real (TSTR) paradigm\cite{vanbreugel2023synthetic}, which evaluates how closely machine
learning models trained on synthetic data resemble models trained on
real data when both are tested on a separate subset of real data. To
clearly illustrate this approach, consider two students studying for the
same exam. If each student studies different materials but achieves
similar results, it can reasonably be inferred that the two sets of
study materials provided comparable knowledge. Similarly, if models
trained on synthetic data produce results similar to models trained on
real data, this suggests that the synthetic data closely replicates the
predictive characteristics of the real data (Fig.~\ref{fig:tstr}).

To practically implement this approach, each synthetic dataset
(generated by Gaussian Copula, CTGAN, and TVAE from SDV; Bayesian
Network, CTGAN, and TVAE from Synthicity) was first scaled and then
split into training subsets. These synthetic training subsets were then
used to train four regression models: XGBRegressor, Random Forest
Regressor, Support Vector Regressor (SVR), and Linear Regression.
Critically, the trained models were tested only on subsets of the real
data, ensuring a valid and realistic evaluation scenario. Multiple
evaluation rounds were conducted using repeated holdout validation, each
time varying the training-test split to obtain robust and statistically
significant results. The predictive performance of these models was
measured using three metrics: Mean Absolute Error (MAE), Mean Squared
Error (MSE), and the Coefficient of Determination (R²). These metrics
were computed both for the models trained on synthetic datasets and for
models trained directly on real data, averaged over multiple evaluation
splits. To facilitate intuitive comparisons, absolute differences
between synthetic-data-trained and real-data-trained model performances
were calculated for each metric, and normalized relative to the
performance of real-data-trained models. These normalized differences
were converted into scores ranging from negative infinity to 1, with a
score of 1 indicating identical performance to the real-data-trained
models. Finally, an Overall Score was derived by averaging these
metric-specific scores, providing an easily interpretable summary of
each synthetic dataset\textquotesingle s predictive utility.

\begin{figure}[htbp]
  \centering
  \includegraphics[width=1\linewidth]{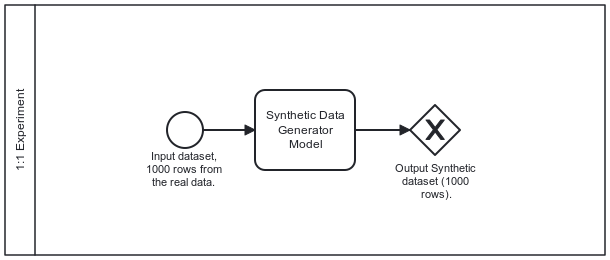}
  \caption{Overview of the 1:1 experimental setting for synthetic data generation: each model is trained on 1,000 real rows and tasked with generating 1,000 synthetic rows.}
  \label{fig:1-1}
\end{figure}

To compare the models under different conditions, two experimental
settings were designed. In the first experiment, all six synthetic data
generators from the two packages were trained on the same subset of
1,000 rows from the real dataset (the energy data described in the
section above). Each model was then asked to generate 1,000 synthetic
rows, resulting in a 1:1 ratio between the input and the output (Fig.~\ref{fig:1-1}).

\begin{figure}[htbp]
  \centering
  \includegraphics[width=1\linewidth]{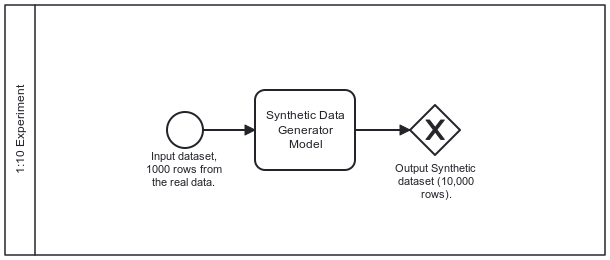}
  \caption{Overview of the 1:10 experimental setting for synthetic data generation: each model is trained on 1,000 real rows and tasked with generating 10,000 synthetic rows.}
  \label{fig:1-10}
\end{figure}

This setup allowed us to evaluate how well each model could replicate the training data without having to extrapolate beyond the scale of the original dataset. In the second experiment, the same synthetic data generators were again trained on the same 1,000 real rows, but this time asked to generate 10,000 synthetic rows, creating a 1:10 ratio between the input and the generated output (Fig.~\ref{fig:1-10}). This second setting was meant to test the models’ ability to scale and generalize, specifically, their ability to generate significantly more data than the original input.

From this process, a total of twelve synthetic datasets were produced:
six from the 1:1 setting and six from the 1:10 setting. All synthetic
datasets were then evaluated using the two previously described metrics:
statistical similarity, which captures how closely the synthetic data
matches the real data in terms of column-wise statistics and
distributions; and predictive utility, which assesses how well machine
learning models trained on synthetic data perform when tested on real
data.

The full implementation and analysis are available in the accompanying
\href{https://github.com/cris1618/syntheticData/blob/main/generate_synthetic_data.ipynb}{GitHub repository}.

\section{Results}
\subsection{Statistical Similarities}

In the 1:1 experiment (described above), six different synthetic
datasets were generated: three from the SDV package and three from the
Synthicity package. Each synthetic dataset was evaluated using the
metric described in the method section, which measures how well the
statistical properties of the synthetic data match those of the original
real dataset on a column-by-column basis. This metric ranges from 0 to
100, with 100 indicating a perfect match, meaning the statistical
properties are identical. In this setting, all models performed well,
with scores ranging between 80 and 96. The best-performing model was the
Bayesian Network from Synthicity, achieving a score of 96.53. The
complete set of results for the 1:1 experiment is shown in Table \ref{tab:1-1}.

\begin{table}[htbp]
\centering
\caption{1:1 Statistical Similarity Scores.}
\label{tab:1-1-similarity}
\begin{tabularx}{\linewidth}{l c c}
\toprule
\textbf{Models} & \textbf{SDV Models} & \textbf{Synthicity Models} \\
\midrule
Gaussian Copula & 91.98 & -- \\
CTGAN           & 80.28 & 88.37 \\
TVAE            & 87.38 & 86.26 \\
Bayesian Network & --   & 96.53 \\
\bottomrule
\end{tabularx}
\label{tab:1-1}
\end{table}

The same evaluation process was applied to the synthetic datasets
generated in the 1:10 experiment. In this case, the similarity scores
were generally lower but still reasonably high, ranging from 70 to just
under 80. Once again, the Bayesian Network from Synthicity was the
best-performing model, achieving a similarity score of 78.25. The
complete results for the 1:10 experiment are presented in Table \ref{tab:1-10}.

\begin{table}[htbp]
\centering
\caption{1:10 Statistical Similarity Scores.}
\label{tab:1-10-similarity}
\begin{tabularx}{\linewidth}{l c c}
\toprule
\textbf{Models} & \textbf{SDV Models} & \textbf{Synthicity Models} \\
\midrule
Gaussian Copula & 77.07 & -- \\
CTGAN           & 71.89 & 75.23 \\
TVAE            & 73.29 & 76.17 \\
Bayesian Network & --   & 78.25 \\
\bottomrule
\end{tabularx}
\label{tab:1-10}
\end{table}

\subsection{Predictive Utility}
The predictive utility of each synthetic dataset was evaluated using the
method described in the previous section. In short, a custom scoring
approach was implemented to quantify how closely the models trained on
synthetic data resembled those trained on real data. This metric ranges
from negative infinity to 1, where a score of 1 indicates that the
synthetic data led to model performance identical to the real data, and
lower values reflect greater divergence. To perform the evaluation, four
regression models, XGBRegressor, Random Forest Regressor, Support Vector
Regressor (SVR), and Linear Regression, were trained separately on both
the real dataset and on each of the twelve synthetic datasets generated
during the 1:1 and 1:10 experiments. For each model, we computed three
standard regression metrics: Mean Absolute Error (MAE), Mean Squared
Error (MSE), and the Coefficient of Determination (R²). The results were
then averaged across the four regression models to ensure robustness.

Table \ref{tab:1-1-ml-performance} presents the averaged model performances for the real dataset
and the synthetic datasets generated in the 1:1 experiment.

\begin{table}[htbp]
\centering
\caption{1:1 ML models average performances.}
\label{tab:1-1-ml-performance}
\begin{tabularx}{\linewidth}{l S[table-format=3.2] S[table-format=5.2] S[table-format=1.2]}
\toprule
\textbf{Dataset for Model Training} & 
\textbf{Avg. MAE} & 
\textbf{Avg. MSE} & 
\textbf{Avg. R\textsuperscript{2}} \\
\midrule
Real Data              & 58.78  & 11649.10 & 0.32  \\
Gaussian Copula (SDV)  & 71.48  & 16400.56 & 0.05  \\
Bayesian Network (SYN) & 57.60  & 11917.78 & 0.31  \\
CTGAN (SDV)            & 118.06 & 23282.25 & -0.36 \\
CTGAN (SYN)            & 96.32  & 18523.87 & -0.08 \\
TVAE (SDV)             & 67.65  & 17335.38 & -0.002 \\
\bottomrule
\end{tabularx}
\end{table}

Table \ref{tab:1-10-ml-performance} shows the corresponding results for the 1:10 experiment.

\begin{table}[htbp]
\centering
\caption{1:10 ML models average performances.}
\label{tab:1-10-ml-performance}
\begin{tabularx}{\linewidth}{l S[table-format=3.2] S[table-format=5.2] S[table-format=1.2]}
\toprule
\textbf{Dataset for Model Training} & 
\textbf{Avg. MAE} & 
\textbf{Avg. MSE} & 
\textbf{Avg. R\textsuperscript{2}} \\
\midrule
Real Data              & 45.69  & 8177.29  & 0.29  \\
Gaussian Copula (SDV)  & 73.91  & 12834.36 & -0.10 \\
Bayesian Network (SYN) & 103.30 & 20567.99 & -0.77 \\
CTGAN (SDV)            & 148.06 & 30546.57 & -1.65 \\
CTGAN (SYN)            & 115.63 & 21322.55 & -0.84 \\
TVAE (SDV)             & 64.97  & 12150.86 & -0.04 \\
\bottomrule
\end{tabularx}
\end{table}

\textbf{Note:} The performance scores on the real data differ between
the 1:1 and 1:10 experiments because different amounts of real data were
used for training. In the 1:1 experiment, only 1,000 rows of real data
were used to train the regression models, matching the size of the
synthetic datasets. In contrast, the 1:10 experiment used 10,000 rows of
real data for training in order to ensure a fair comparison with the
larger synthetic datasets generated in that setting.

In the 1:1 experiment, the synthetic datasets generated by the six
synthetic data generators from the two packages performed reasonably
well overall in terms of predictive utility (as measured by the custom
score ranging from negative infinity to 1, described in the method
section). The Bayesian Network from Synthicity achieved the highest
score by far (0.97), followed by the Gaussian Copula from SDV (0.51). A
score of 0.97 indicates that the performance of the regression models
trained on the synthetic data generated by the Bayesian Network was
nearly identical to the performance of the same models trained on the
real data. This suggests that the synthetic data closely matched the
underlying structure and distribution of the real dataset, an ideal
outcome for synthetic data generation. The complete results for the 1:1
experiment are presented in Table \ref{tab:1-1-predictive-utility}.

\begin{table}[htbp]
\centering
\caption{1:1 Predictive Utility Scores.}
\label{tab:1-1-predictive-utility}
\begin{tabularx}{\linewidth}{l S[table-format=1.2] S[table-format=1.2]}
\toprule
\textbf{Models} & 
\textbf{SDV Models} & 
\textbf{Synthicity Models} \\
\midrule
Gaussian Copula (SDV)   & 0.51  & \text{--}    \\
CTGAN                   & -0.37 & 0.17  \\
TVAE                    & 0.45  & 0.45  \\
Bayesian Network (SYN)  & \text{--}   & 0.97  \\
\bottomrule
\end{tabularx}
\end{table}

The results were notably different in the 1:10 experiment. In this
setting, overall performance dropped significantly, with predictive
utility scores considerably lower than those observed in the 1:1
experiment. Moreover, the models that performed best in the 1:1 setting
did not necessarily perform well here. In this case, the best-performing
model was TVAE from SDV, which achieved a predictive utility score of
0.31, followed by the Gaussian Copula from SDV with a score of 0.15. All
other models produced negative scores, indicating a substantial drop in
performance and suggesting that the synthetic data they generated
diverged significantly from the real data, an undesirable outcome in the
context of synthetic data generation. The full set of results for the
1:10 experiment is shown in Table \ref{tab:1-10-predictive-utility}.

\begin{table}[htbp]
\centering
\caption{1:10 Predictive Utility Scores.}
\label{tab:1-10-predictive-utility}
\begin{tabularx}{\linewidth}{l S[table-format=1.2] S[table-format=1.2]}
\toprule
\textbf{Models} & 
\textbf{SDV Models} & 
\textbf{Synthicity Models} \\
\midrule
Gaussian Copula (SDV)   & 0.15   & \text{--}     \\
CTGAN                   & -2.85  & -1.33  \\
TVAE                    & 0.31   & -0.54  \\
Bayesian Network (SYN)  & \text{--}     & -1.13  \\
\bottomrule
\end{tabularx}
\end{table}

\section{Discussion \& Conclusion}
This study set out to compare synthetic data generators from two of the
most prominent open-source Python packages: SDV and Synthicity. Our goal
was to evaluate both usability and performance, and to determine whether
one package consistently outperforms the other in generating
high-quality synthetic data. Based on the results, there is no clear
overall winner in terms of performance. When it comes to statistical
similarity, all models from both packages performed reasonably well,
with no significant distinction between the two. In terms of predictive
utility, however, the Bayesian Network model from Synthicity clearly
outperformed all others in the 1:1 experiment. Notably, SDV does not
include an implementation of a Bayesian Network, which may explain part
of the difference, especially considering that the structure of Bayesian
Networks can be particularly effective for capturing dependencies in
datasets like the one used in this study\cite{batbaatar2018deepenergy} (i.e., household
energy consumption, as described in the method section).

Beyond raw performance, another important consideration is usability,
including documentation quality and user experience. In this regard, SDV
clearly stands out. It offers excellent documentation, a larger and more
active community, and comprehensive tutorials that make the framework
accessible, even to researchers and users with limited programming
experience. The process of generating synthetic data with SDV is
straightforward and well-supported. In contrast, Synthicity's
documentation is less complete and can be difficult to follow. During
the course of this study, we encountered issues using Synthicity,
particularly with larger datasets, and the troubleshooting process was
not well-supported by the existing documentation. In conclusion, while
both packages offer competitive performance, SDV excels in usability and
support, making it a more user-friendly option. However, for users
specifically interested in models like Bayesian Networks, Synthicity may
be a better choice.

One of the main challenges in synthetic data generation lies in
producing large amounts of realistic data from a small real dataset. The
1:1 and 1:10 experiments were specifically designed to explore this
challenge. While the 1:1 scenario tests how well a generator can
replicate what it has seen, the 1:10 scenario evaluates how well it can
extend and generalize, an essential capability in low-data settings. The
results clearly show that synthetic data generators performed
significantly better in the 1:1 case. In this setting, most models
achieved high statistical similarity scores, and predictive utility
remained relatively strong, with the Bayesian Network from Synthicity
reaching a near-perfect score of 0.97. This suggests that when the
output size matches the input, generators are more capable of preserving
the underlying data structure. In contrast, the 1:10 experiment revealed
a marked drop in performance. While statistical similarity scores
remained acceptable (between 70 and 80), predictive utility suffered
noticeably. Only TVAE from SDV achieved a moderately positive score
(0.31), while most other models, including the same ones that performed
well in the 1:1 case, yielded negative utility scores. These results
highlight the limitations of current synthetic data generators when
tasked with generalizing beyond their input size. They also underscore
the need for more robust generative models capable of scaling synthetic
datasets without sacrificing predictive quality, a critical requirement
in real-world scenarios where large datasets are needed but only small
samples are available.

Another notable result that deserves further exploration is the observed
performance disparity between statistical and deep learning-based
methods across the two experimental settings. In the 1:1 experiment,
statistical models such as Gaussian Copula and Bayesian Network
outperformed deep learning-based models like CTGAN and TVAE in terms of
predictive utility. This superior performance can be attributed to the
nature of statistical models, which rely on explicit probabilistic
assumptions and are well-suited for capturing the underlying
distributions in smaller datasets. For instance, Bayesian Networks have
demonstrated effectiveness in modeling complex dependencies in building
energy consumption data, providing accurate predictions even with
limited data\cite{geraldi2019bayesian}.

Conversely, in the 1:10 experiment, where the synthetic data generators
were tasked with producing ten times more data than the original input,
deep learning-based models exhibited better performance. Specifically,
TVAE from SDV achieved the highest predictive utility score among all
models in this setting. Deep learning models like TVAE are designed to
capture intricate, high-dimensional patterns in data, making them more
adept at generalizing from limited inputs to generate larger synthetic
datasets. However, it\textquotesingle s important to note that while
TVAE outperformed other models in the 1:10 experiment, its predictive
utility score was still lower compared to the top-performing models in
the 1:1 experiment, indicating challenges in maintaining data quality
when scaling up synthetic data generation. Interestingly, CTGAN, another
deep learning-based model, consistently underperformed in both
experimental settings. This consistent underperformance suggests that
CTGAN may have limitations in capturing the complex dependencies present
in the energy consumption dataset used in this study. Previous research
has also indicated that CTGAN\textquotesingle s performance can vary
depending on the dataset characteristics and that it may not always be
the most effective choice for tabular data generation\cite{synthcity2023github}.

These findings suggest that the choice between statistical and deep
learning-based synthetic data generation methods should consider the
specific requirements of the task at hand, including the size of the
available real dataset and the desired scale of the synthetic data.
Statistical models may be more appropriate for scenarios with limited
data and a need for high fidelity in replication, while deep learning
models might be better suited for applications requiring the generation
of larger synthetic datasets, albeit with potential trade-offs in data
quality.

Finally, it\textquotesingle s important to address why models with the
same name, CTGAN and TVAE, showed different performances when
implemented in the SDV and Synthicity packages. Although these models
are conceptually based on the same architectures, the way each package
implements them differs in key ways that can impact performance. For
example, Synthicity uses deeper neural networks with more parameters and
applies stronger regularization methods like dropout and weight decay,
which can help prevent overfitting but may require more training data to
work effectively {[}42{]}. In contrast, SDV's versions tend to use
smaller, simpler networks and rely on techniques like mode-specific
normalization and PacGAN to improve learning from small datasets.
Another factor is the training process itself. SDV uses a fixed number
of training epochs with relatively conservative learning rates, while
Synthicity uses more dynamic training strategies with longer training
durations and early stopping. These choices can make
Synthicity\textquotesingle s models more flexible but also more
sensitive to the nature of the input data. Furthermore, the way data is
preprocessed before being fed into the models varies between the two
frameworks. SDV uses its own transformation toolkit (RDT), which handles
continuous and categorical columns differently from Synthicity's
internal preprocessing methods\cite{sdv2024docs}.

Together, these differences help explain why TVAE from SDV was the
best-performing deep learning model in the 1:10 experiment, while its
counterpart from Synthicity did not achieve the same level of utility.
Likewise, CTGAN, despite being available in both packages, performed
poorly across both settings, possibly due to its sensitivity to
architectural and training hyperparameters, which differ significantly
between the two environments\cite{akiya2024comparison}.

\bibliographystyle{unsrt}
\bibliography{ref}

\end{document}